# Dyadformer: A Multi-modal Transformer for Long-Range Modeling of Dyadic Interactions

David Curto[*,1,2], Albert Clapés[*,3,4], Javier Selva[*,1,3], Sorina Smeureanu[1,3],
Julio C. S. Jacques Junior[3], David Gallardo-Pujol[1], Georgina Guilera[1], David Leiva[1],
Thomas B. Moeslund[4], Sergio Escalera[1,3,4], Cristina Palmero[1,3]

[1]Universitat de Barcelona [2]Universitat Politècnica de Catalunya
[3]Computer Vision Center [4]Aalborg University
david.curto@estudiantat.upc.edu, alcl@create.aau.dk, jaselvaca@ub.edu,
{crpalmec7, ssmeursm28}@alumnes.ub.edu, jjacques@cvc.uab.cat,
{david.gallardo, gguilera, dleivaur}@ub.edu, tbm@create.aau.dk, sergio@maia.ub.es

## Abstract

*Personality computing has become an emerging topic in computer vision, due to the wide range of applications it can be used for. However, most works on the topic have focused on analyzing the individual, even when applied to interaction scenarios, and for short periods of time. To address these limitations, we present the Dyadformer, a novel multi-modal multi-subject Transformer architecture to model individual and interpersonal features in dyadic interactions using variable time windows, thus allowing the capture of long-term interdependencies. Our proposed cross-subject layer allows the network to explicitly model interactions among subjects through attentional operations. This proof-of-concept approach shows how multi-modality and joint modeling of both interactants for longer periods of time helps to predict individual attributes. With Dyadformer, we improve state-of-the-art self-reported personality inference results on individual subjects on the UDIVA v0.5 dataset.*

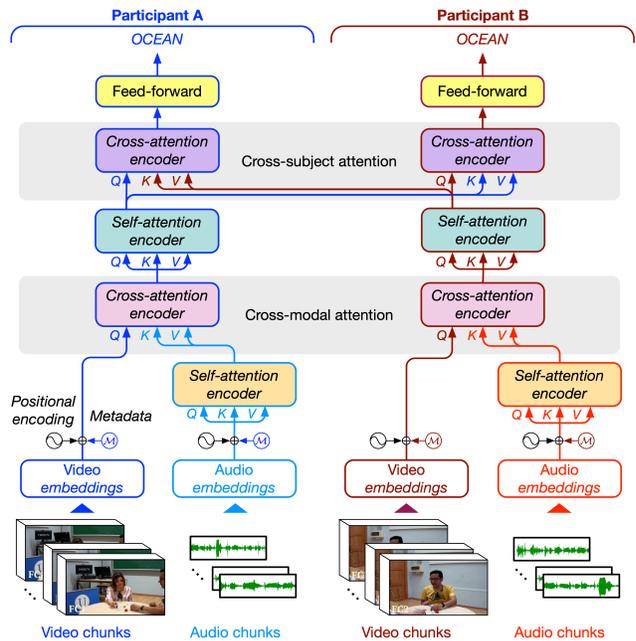

Figure 1. Proposed Dyadformer including different kinds of attention (self, cross-modal, and cross-subject). The model jointly infers the self-reported personality of both participants (A and B). Model complexity is reduced by sharing weights between parallel encoders (as illustrated by their colors) and across layers within each encoder. $\mathcal{M}$ are the corresponding metadata embeddings of each participant added to both their video and audio embeddings.

## 1. Introduction

In the past years, human interaction and, in particular, dyadic interaction, have been deeply studied by both psychological and artificial intelligence research communities [9, 8, 17]. This rising trend has led to remarkable development not only in data collection [10, 6, 12], but also in defining methods for automatically understanding and modeling interpersonal social signals [63, 52]. The way people adapt and react to such signals and behaviors during a conversation depends not only on their individual characteristics (*e.g.*, personality) but also on the specific situation and their shared history [9]. For example, one might behave more relaxed during a conversation with a friend than in a meeting with their foreman. When analyzing social interactions from a computational perspective, all these influential factors should be taken into consideration to truly understand human behavior, even when the focus is on predicting individual attributes such as personality traits [72]. However, this is still not the norm throughout the literature [58].

---
[*]These authors contributed equally to this work.

Despite the growing interest in this area, current computational approaches for social interaction understanding present some other shortcomings. For instance, long-term modeling is crucial in interaction settings, as more complex dynamics emerge at different time scales, and an event may unchain effects that take time to be observed [9]. In the case of personality computing in such scenarios, the need for long-term modeling is heightened, as behavioral manifestations of certain traits may not be fully observed in short periods of time. Hence, more time is needed to find salient patterns arising during the interaction that can be associated to given traits [24]. Most existing works that deal with longer-term modeling have generally been addressed through single frame descriptors averaged over whole sequences [46], missing to represent the temporal evolution of features. Another aspect that fails to be properly modeled when assessing individual attributes in dyadic interactions is the joint modeling of both interlocutors. Despite its importance for triggering individual behaviors that provide insights on individual features [4], most of the works that do model it are focused on analyzing interaction attributes.

In this work, we propose a novel architecture to leverage long-term information for joint modeling of both interlocutors in dyadic scenarios. More precisely, we predict the personalities of both interactants by considering not only the audio-visual information and contextual factors (referred to as metadata) independently for each one but also by explicitly modeling their interaction. The proposed model, *Dyadformer*, mainly consists of two stages: (1) a *cross-modal* stage where cross-attention encoders fuse multi-modal information, and (2) a *cross-subject* stage which aims to shape the interaction by performing double cross-attention (see Fig. 1). Our contributions are summarized as follows:

- To our knowledge, this method is the first one to jointly model (and infer) self-reported personality in dyadic interactions using time windows of up to ∼30 seconds.

- Inspired by the classical decoder block of the Transformer network [69], we leverage a cross-attention mechanism to both fuse modalities and allow information to flow between subjects.

- Dyadformer obtained state-of-the-art results on the large-scale UDIVA v0.5 [58] dataset, by reducing previous participant-level error by a 12.5% (from 0.812 to 0.722) when predicting self-reported personality.

## 2. Related Work

**Social signals and behaviors in context.** Dyadic and small group interactions are a rich source of overt behavioral cues. They can provide insight into our personal attributes and cognitive/affective inner states dependent upon the context in which they are situated. Context can take many forms, from the interaction partner's attributes and behaviors to spatio-temporal and multi-view information. Joint modeling of both interlocutors and/or other sources of context have been extensively considered when trying to measure interpersonal constructs [13, 82], individual social behaviors [14, 81] and impressions [81, 59], and even empathy [59]. When considering individual attributes instead, context has often been misrepresented, in spite of extensive claims on its importance [4, 74, 70, 56].

In recent years, interlocutor-aware approaches have started to gain more attention, especially for emotion recognition in conversation [60, 50]. Richer contexts have been captured by explicitly modeling the temporal dimension, which was traditionally done through recurrent approaches [53, 77, 26]. However, recent works have started using BERT/Transformer-like architectures [83, 49]. Beyond text, using additional modalities has also been proposed, e.g., raw audiovisual data [79, 31, 76, 36], or speech cues and personality of the target speaker [47]. Regarding context-aware personality recognition (the focus of this work), a similar trend is seen, but the literature is even scarcer, as discussed next.

**Automatic personality recognition.** Personality is defined as the manifestation of individual nuances in patterns of thought, feeling, and behavior, that remain relatively stable over time [65]. In the personality computing field [71], it is usually characterized by the basic Big Five traits [30] (*Open-mindedness*, *Conscientiousness*, *Extraversion*, *Agreeableness*, and *Negative emotionality*), often referred to as OCEAN, based on self-reported assessments. Most works focus on personality recognition from the individual point of view [32, 75], even in a dyadic or small group conversational context [1], using only features from the target person. Initial studies tended to use handcrafted features from gestures and speech [57], while more recent works rely on deep learning approaches from raw data [55].

Few methods propose interlocutor- or context-aware methods for self-reported personality recognition in small group interactions. Most recent works focus on personality analysis on social media, generally limited to the textual modality (see [54] for a complete review), involving much more people while missing useful cues from face-to-face interactions. The work of [66] was one of the firsts to model conversation in small group settings, leveraging turn-taking temporal evolution from transcript features but focusing on apparent personality recognition (i.e., personality reported by external observers [38]). Other works have also focused on modeling transcribed interviews [35], but disregarding the interviewer. In [18], authors regressed individual and dyadic features of personality and social impressions utilizing handcrafted descriptors of prosody, speech, and visual activity. [51] proposed an interlocutor-modulated recurrent attention model with turn-based acoustic features. [80] predicted personality and performance labels by correlation analysis of co-occurrent key action events, which were extracted from head and hand pose, gaze, and motion intensity features. The use of person metadata (*e.g.*, gender, age, ethnicity, and perceived attractiveness) together with audiovisual data has only been applied in [61]. However, their goal was to better approximate the crowd biases for apparent personality inference in one-person videos.

**Long-term modeling in personality computing.** The need for longer-term modeling in personality regression tasks is highlighted in [64]. The authors proposed a model based on facial features for individual apparent and self-reported personality, but limited to 3-second time windows. Others have attempted long-term modeling of features for personality inference, but most are limited to compute sequence representations by averaging small clips or individual frames features [46, 40], which miss temporal relationships. As far as we know, only one previous work has used up to 1 minute without aggregating across clips [67]. However, they focused on first-impressions regression, which does not benefit from longer temporal windows. In the past years, a new family of architectures has risen to address some limitations of traditional recurrent methods [41], i.e., the Transformer [69]. Originally designed for machine translation, it has shown impressive results for many sequence modeling tasks in a plethora of modalities [15, 16, 5]. These models are capable of attending to long-range data dependencies with few layers, allowing them to learn very useful representations. Recently, some works have started using Transformer-like architectures to model personality. However, these works tend to focus on the apparent personality of individuals alone [29] by only modeling text features, generally on social media posts [45, 78]. We focus on self-reported personality on real face-to-face dyadic interactions, which our proposed architecture explicitly exploits, and use the Transformer to model video, audio, and metadata modalities altogether.

In this line, [58] is, to the best of our knowledge, the only work on self-reported personality regression in dyadic scenarios that takes multi-modal context into account. However, this approach has some limitations. First, participant personality was regressed from just 3-second chunks, which may not be enough to properly model long-term interactions. In this work, we input longer clips (up to 30 seconds), allowing the model to learn useful longer-range relationships. Second, multi-modal fusion was done simply by concatenating the information from the video and audio modalities. In this work, we leverage multi-modal Transformers that exploit useful features from each source by looking at interdependencies, and fuse them in a shared representation space. Finally, despite [58] combining information from both individuals in the interaction, only the personality of the target subject was regressed. Our Dyadformer explicitly models the behavior from both individuals simultaneously, through our proposed two-stream cross-attentional Transformer, to eventually predict their personalities jointly.

**Multi-modal Transformers.** Our work is related to the recent use of Transformers [69] to learn multi-modal representations. The most common approach employs contrastive losses to bring paired samples (such as video and caption [27] or subtitles [48]) closer together. This is generally used for captioning or retrieval tasks, where both modalities provide similar information and the aim is to *translate* between them. However, in our setting we expect audio and video to convey different complementary information, for which we explore two better suited Transformer families. The first one uses a BERT-like [15] stream which concatenates modalities along the temporal dimension [44, 21] before input, effectively doubling sequence length. Nevertheless, as Transformers scale quadratically with input length, these methods incur in memory efficiency limitations. The second one solves this by using separate cross-attention streams [39, 37], replacing self-attention to allow both modalities to attend and enrich each other (see Sec. 3.2 for details), while the separate streams allow for independent modeling and maintain sequence length. This design has generally been used to fuse two modalities, as is our case, but it can easily be extended further [84]. In preliminary experiments, we tested using a BERT-like approach but, when compared with the latter alternative and in our setting, we found it to underperform. For this reason, we opted for cross-attentional streams to fuse multi-modal information and go one step further by also using this technique to model cross-subject interactionss.

## 3. Methodology

In this section, we present the Dyadformer ( Fig. 1), composed of a set of *attentional encoder modules*. Each of these is a stack of *Transformer layers* [69]. A complete transformer layer is composed of two or more *sub-layers*. Each sub-layer executes a core *block*, followed by a residual connection and layer normalization [2]. We define these four elements in Sec. 3.1, and describe the cross-modal and cross-subject attention that form the full architecture in Sec. 3.2.

### 3.1. The Transformer

The core of the Transformer layer is a non-local operation [73], which allows every element in the input sequence to access information from any other. This is achieved through a special form of attention. To compute it, the input representation $J \in \mathbb{R}^{T \times d_\text{w}}$ is mapped to a set of queries $Q \in \mathbb{R}^{T \times d_\text{k}}$, and a memory $M \in \mathbb{R}^{T \times d_\text{w}}$ is mapped to a set of paired keys $K \in \mathbb{R}^{T \times d_\text{k}}$ and values $V \in \mathbb{R}^{T \times d_\text{k}}$, where $d_\text{k} = d_\text{w}/h$ and $h$ is the number of heads (defined below). In the Transformer, the non-local operation is instantiated as the dot-product between $Q$ and $K$ in order to generate an affinity (attention) matrix that weights how much each value should contribute to the augmented representation of every other value. In general, $J$ will be equal to $M$, hence this attention is called self-attention. The output of the self-attention operation is computed as

$$\text{Att}(Q, K, V) = \text{softmax}(\frac{QK^T}{\sqrt{d_\text{k}}})V. \quad (1)$$

However, in order to build a full transformer layer, attention is not all you need. A complete transformer layer is composed by sub-layers that can be defined as $\text{SubLayer}^n_\text{Block}(x) = \text{LayerNorm}(x + \text{Block}(x))$, where the output of one sub-layer is fed as input to the next, $n$ references the attentional encoder module that the sub-layer belongs to and $Block$ will either be Multi-Head self-Attention (MHA) or *position-wise Feed-Forward Network* (FFN) [69] which we define next.

**Sub-Layer blocks.** In order to allow for the model to attend to different information in a single sub-layer, [69] proposed the Multi-Head self-Attention (MHA). Similar to the multiple filters of a single convolutional layer, the multi-head self-attention maps $J$ and $M$ to $h$ different representation sub-spaces to perform different attention operations. Then, the output of each head is concatenated and mapped back to a common $d_\text{w}$-dimensional representation through a linear transformation $W^O \in \mathbb{R}^{(h*d_\text{k}) \times d_\text{w}}$. More formally,

$$\text{MHA}^n_f(J, M) = \text{Concat}(\text{head}_1, ..., \text{head}_h)W^O,$$
$$\text{where head}_i = \text{Att}(JW^Q_{n,f,i}, MW^K_{n,f,i}, MW^V_{n,f,i}), \quad (2)$$

$W^\cdot_{n,f,i} \in \mathbb{R}^{d_\text{w} \times d_\text{k}}$ and $f$ is the sub-layer within which the MHA block is, detailed next.

In practice, every sub-layer in a Transformer layer contains a MHA block except for the last one, which contains a FFN block. It is composed of two linear layers with GELU [33] activation function in between, i.e., $\text{FFN}(x) = \text{GELU}(xW_{F1})W_{F2}$, where $x \in \mathbb{R}^{d_\text{w}}$ is the embedding corresponding to one timestep, and $W_{F1} \in \mathbb{R}^{d_\text{w} \times (4*d_\text{w})}$ and $W_{F2} \in \mathbb{R}^{(4*d_\text{w}) \times d_\text{w}}$ are matrices of weights. In practice, point-wise FFN are equivalent to applying a fully connected (FC) layer repeatedly and independently to each timestep (these can also be seen as two 1D convolutional layers with kernel size 1). Note that FFN layers are also dependent on $n$ and $f$, but we have omitted denoting those for simplicity.

**Attentional Encoder modules.** In our design, we use two main modules to build the complete architecture: the self-attention encoder $\text{SA}(J)$, which is used to enhance features by attending to themselves, and the cross-attention encoder $\text{CA}(J, M)$, which is used to allow for a set of features to attend to a different source. The former is composed of Transformer layers with a single $\text{SubLayer}^\text{SA}_\text{MHA}(J)$, while the latter is composed by two, $\hat{J}_l = \text{SubLayer}^\text{CA}_{\text{MHA}_1}(J_l)$ and $\text{SubLayer}^\text{CA}_{\text{MHA}_2}(\hat{J}_l, M)$. As mentioned earlier, in both cases ($\text{SA}(J)$ and $\text{CA}(J, M)$), the described sub-layers are followed by a last $\text{SubLayer}^n_\text{FFN}$ sub-layer. It is important to note that, in the cross-attention encoder, while the input $J_l$ of layer $l$ is the output from the previous cross-encoder layer, $M$ is the same for all layers. This allows $J_l$ features to iteratively attend to $M$ to be progressively augmented.

**Positional Encoding.** Finally, as the self-attention operation is agnostic to relative position among input elements, [69] proposed using positional encodings to indicate the order of the input sequence by a composition of sine and cosine functions at varying frequencies. We followed the same procedure to indicate the ordering of the timesteps in the sequences from both input modalities.

### 3.2. The Dyadformer

Dyadformer (depicted in Fig. 1) is a multi-subject multi-modal architecture that follows the aforementioned transformer layers. The Dyadformer receives as input a sequence of $T$ small, consecutive and temporally aligned video/audio chunks and infers the personality traits for both subjects in a dyadic interaction. It is composed of two main streams, each of which simultaneously processes a single subject.

As discussed in Sec. 2, context and interpersonal features are crucial to predict individual features in dyadic and small group interaction scenarios. For this reason, we propose a model which is capable of (a) fusing information from multiple sources (video, audio, and contextual metadata), and (b) allowing per-subject streams to access each other, in order to consider crossed influence during the interaction. To satisfy both, we go beyond self-attention, where $J = M$, and also use cross-attention, where $J \neq M$. Cross-attention works similarly to encoder-decoder attention in [69], where the input and memory come from different sources. For a transformer focusing on dyadic interactions, the target ($J$) will be from the subject of interest, while the memory ($M$) will be from the other one. The intuition behind this is to allow information from a given subject to *query* for useful information from the other. But first, each stream will create an individual representation for each subject. In order to do so, we draw inspiration from multi-modal transformer models [84, 39, 37], and use this same cross-attentional mechanism to fuse data coming from video and audio modalities. In this cross-modal module, $J$ is from the video modality, while $M$ is from the audio one, thus enriching video infor-mation with the audio signal. Finally, personality scores for both individuals are predicted jointly.

**Input.** We temporally divide videos and audios into small chunks first. Next, we precompute per-chunk feature representations using pre-trained networks (see Sec. 4.1 for details). Doing so, we can then feed our model with two pairs of sequences $(X^p, U^p)$, where $p \in \{A, B\}$ denote the participants, $X^p = [x_1^p, \ldots, x_T^p]$ is a sequence of precomputed video features and $U^p = [u_1^p, \ldots, u_T^p]$ is the corresponding sequence of precomputed audio features. Note that $X^A$, $U^A$, $X^B$, and $U^B$ are all temporally aligned. Apart from these, the model also receives the metadata handcrafted features, namely $m^p$. Then, the precomputed video and audio features, as well as metadata, are linearly projected into $d_w$-dimensional embeddings via three independent linear layers. Next, for each participant, positional encodings and their respective metadata embeddings are summed to their video and audio embeddings. Given $m^p$ has no temporal dimension, before the summation, $m^p$ is replicated $T$ times using the outer product operation: $\mathcal{M} = \mathbf{1} \otimes m^p$, where $\mathbf{1}$ is a $T$-sized vector of ones.

**Cross-modal and cross-subject attention.** In order to build the multi-modal representation for each subject, we first feed the audio features $U^p$ to an audio encoder module $\text{SA}_{\text{aud}}$ composed by $L_{\text{aud}}$ layers, such that $\hat{U}^p = \text{SA}_{\text{aud}}(U^p)$ where $\hat{U}^p \in \mathbb{R}^{T \times d_w}$. Then, we use a cross-encoder $\text{CA}_{\text{vid}}$ with $L_{\text{xm}}$ layers to enhance video features $X^p$ with the new audio features, such that $\hat{X}^p = \text{CA}_{\text{vid}}(X^p, \hat{U}^p)$, where $\hat{X}^p \in \mathbb{R}^{T \times d_w}$.

The enhanced video features of each subject $\hat{X}^p$ are transformed through a subject encoder $\text{SA}_{\text{sbj}}$ with $L_{\text{sbj}}$ layers, such that $S^p = \text{SA}_{\text{sbj}}(\hat{X}^p)$, in order to learn rich relationships within individual subject features. This subject encoder is followed by a cross-encoder with $L_{\text{xs}}$ layers as to allow the features from each subject to draw relevant information from each other, such that $\hat{S}^A = \text{CA}_{\text{sbj}}(S^A, S^B)$ and $\hat{S}^B = \text{CA}_{\text{sbj}}(S^B, S^A)$, where $S^p$ and $\hat{S}^p \in \mathbb{R}^{T \times d_w}$.

**Inference.** For a given sequence, to infer the personality of the participant $p$ in the dyad, we feed the output subject representations $\hat{S}^p = \{\hat{s}_1^p, \ldots, \hat{s}_T^p\}$ through an average pooling and two FC layers in order to regress the final OCEAN values for $p$, i.e. $\hat{o}^p = ((z^p W_{\text{FC1}}) W_{\text{FC2}})$, where $z^p = \frac{1}{T} \sum_{t=1}^{T} \hat{s}_t^p$, $z^p \in \mathbb{R}^{d_w}$, $W_{\text{FC1}} \in \mathbb{R}^{d_w \times 4*d_w}$ and $W_{\text{FC2}} \in \mathbb{R}^{4*d_w \times 5}$.

## 4. Experimental evaluation

Next, we experimentally evaluate a set of variants of the Dyadformer architecture for the task of self-reported personality traits regression and discuss the obtained results.

### 4.1. Data

**UDIVA v0.5 dataset.** All the experiments were based on UDIVA v0.5[1], a preliminary subset of the UDIVA dataset [58]. This subset is a highly varied multi-modal, multi-view dataset of zero- and previous-acquaintance, face-to-face dyadic interactions. It consists of 145 interaction sessions, where 134 participants (17-75 years old, 55.2% male) arranged in dyads performed a set of four tasks in a lab setting: *Talk*, *Lego*, *Ghost*, and *Animals*.

Contained speech data is multi-lingual, with 73.1% of the sessions in Spanish, 17.25% in Catalan, and 9.65% in English. The UDIVA v0.5 dataset provides one frontal camera view per participant, the audio streams from each participant's lapel microphone, and participants' and sessions' metadata, as well as other modalities and annotations. Personality trait ground truth values were obtained from a self-reported BFI-2 questionnaire [22], and are provided as z-scores.

In our experiments, we used the audio-visual data and the metadata from [58]. The latter consists of 21 features encoding information of an individual (age, gender, cultural background, session index of participant, and pre-session mood/fatigue), session (task order within the session and its difficulty), and dyadic information (relationship between interactants). The dataset is divided into three subject-independent splits: 116/18/11 sessions and 99/20/15 participants in training, validation, and test, respectively.

**Pre-segmented chunks and feature extraction.** For the sake of comparison, we utilized the same set of video and audio chunks used in [58], provided by the authors. In their work, chunk availability was limited by a face detection algorithm, such that chunks with no detected face were discarded. Given also the difference in duration throughout sessions and tasks, the final number of chunks per task was uniformly subsampled based on the median. Then, to sample contiguous sequences of $T$ chunks, some of them have been further discarded. Given these limitations, some tasks do not contain many chunks and, to avoid losing more data, we limited our experiments to $T \leq 12$.

After all, we end up with a substantially smaller dataset: resulting train, validation and test splits contain, respectively, 94,960/15,350/7,870 pre-segmented chunks (equivalent to 67.5/10.9/5.6 hours). As Transformers are known to be data hungry [16], we follow other works [28, 42] who have successfully trained Transformers on smaller datasets by leveraging backbones pre-trained on Kinetics [11].

Each video chunk is composed of 32 frames at 12.5 fps (∼2.56 seconds) at a spatial resolution of $224 \times 224$ pixels (normalized between $[0, 1]$), whereas each audio chunk was 3 seconds long, acquired at 44.1 kHz, and time-synchronized to its respective video chunk (i.e., the centers of corresponding video/audio chunks are aligned). Video, audio, and contextual metadata features are generated for each subject individually. Visual features are computed with R(2+1)D [68] pre-trained on IG-65M [25] and Kinetics [11]. We also fine-tuned its 5th block on the training set of UDIVA v0.5 during 13 epochs (after having replaced the last fully connected layer by another one of size 5 to predict OCEAN). Once trained, all the pre-segmented chunks of UDIVA v0.5 were reprocessed and the 512-dimensional feature representations output by the second to last layer of R2+1D were saved. Analogously, for audio, we used a VGGish [34] pre-trained on AudioSet [23] to compute a 128-dimensional representation for each audio chunk. Sequences of $T$ such video/audio precomputed features were used as input for each subject in our method.

---

[1] https://chalearnlap.cvc.uab.es/dataset/41/description/.

### 4.2. Parameters and implementation details

Following [15], we fixed $d_w = 768$ and $h = 12$, and hence $d_k = 64$. We set $L_{aud} = L_{xm}$ and $L_{sbj} = L_{xs}$ for our experiments. To maximize the number of consecutive $T$-length training sequences, they were sampled with a stride of 1 chunk. Metadata was included for all the experiments if not otherwise stated, based on the findings of [58].

Transformer models quickly grow in number of parameters. In our simplest model (see $TF_v$ in Tab. 1) one Transformer layer accounted for ~7.1M, whereas 8 layers accounted for ~56.8M parameters (disregarding the backbones and final linear layers). Nevertheless, recent studies on Transformer models in NLP [3, 43], later extended to the audio-visual domain with similar results [44], have shown that weight sharing does not hurt representational power nor performance, while allowing for lighter and faster-to-train models. For this reason, in this work we always shared weights between all equivalent layers of both subject's streams. In other words, both streams were exactly the same. Also, for experiments where layers for any given module $L. \geq 1$, we shared parameters across them (*e.g.*, all cross-modal Transformer layers share weights).

Our model was trained by minimizing a MSE loss measuring the error of the inferred personality traits at sequence level versus its associated ground truth: $\mathcal{L} = \sum_{p \in \mathcal{P}} \sum_{i=0}^{5} (o_i^p - \hat{o}_i^p)^2$, $o^p$ is the ground truth of self-reported personality and $\mathcal{P} \subseteq \{A, B\}$ (depending on the experiment). Model weights were trained by minimizing $\mathcal{L}$ via SGD optimization with weight decay $5e^{-3}$. Training was early stopped after 6 epochs if no improvement was observed on the validation loss. The learning rate was initially set to $5e^{-4}$ and reduced by a factor of 2 after 3 epochs without improvement. The dropout rate throughout all the layers in the architecture was set to 0.2.

| Arch. | $L$ | | MSE$_{seq}$ | | MSE$_{part}$ | | Params |
|---|---|---|---|---|---|---|---|
| | | | $T=6$ | $T=12$ | $T=6$ | $T=12$ | |
| TF$_v$ | 2 | | 0.807 | 0.771 | 0.742 | 0.732 | 10.0M |
| | 4 | | 0.857 | 0.792 | 0.781 | 0.744 | |
| | 6 | | 0.919 | 0.856 | 0.837 | 0.807 | |
| | 8 | | 0.948 | 0.860 | 0.867 | 0.804 | |
| | $L_{xm}$ | $L_{xs}$ | $T=6$ | $T=12$ | $T=6$ | $T=12$ | |
| DF$_{xm}$ | 1 | - | **0.797** | 0.767 | **0.738** | 0.732 | 19.4M |
| | 2 | - | 0.845 | 0.767 | 0.777 | **0.722** | |
| | 3 | - | 0.880 | 0.802 | 0.824 | 0.762 | |
| DF$_{xs}$ | - | 1 | 0.802 | 0.768 | 0.763 | 0.745 | 19.4M |
| | - | 2 | 0.831 | 0.760 | 0.778 | 0.738 | |
| | - | 3 | 0.843 | 0.767 | 0.794 | 0.743 | |
| DF$_{xm,xs}$ | 1 | 1 | 0.831 | 0.760 | 0.794 | 0.741 | 36.0M |
| | 1 | 2 | 0.847 | 0.765 | 0.802 | 0.748 | |
| | 2 | 1 | 0.854 | **0.738** | 0.809 | **0.722** | |
| | 2 | 2 | 0.894 | 0.758 | 0.842 | 0.737 | |

Table 1. Ablation of different architectures and sequence lengths ($T$ chunks) in terms of average sequence- and participant-level mean squared errors: TF$_v$, a Transformer on each subject's sequence separately; DF$_{xm}$ or DF$_{xs}$, the Dyadformer with only cross-modal ("xm") or cross-subject ("xs") attention respectively; and DF$_{xm,xs}$ with both. $L.$ are the number of layers in the encoders. Best result per column in bold.

### 4.3. Evaluation metrics

For the following experiments, we report the average per-trait Mean Squared Error (MSE) at two levels: (a) *sequence-level* (MSE$_{seq}$), where the error was computed for every $T$-length sequence by comparing the predictions against ground truth personality of the subject appearing in them. The MSE$_{seq}$ reported is the mean over all the $T$-length sequences in the test set; and (b) *participant-level* (MSE$_{part}$), for which we first aggregated the predictions over all the sequences of a given participant by the median, and then compared it to that participant's personality ground truth. In contrast to MSE$_{seq}$, MSE$_{part}$ removes bias towards participants that appear more in the test set, hence being a more balanced metric for this problem. We choose to report both in this work to compare the effect of the different aggregation mechanisms.

### 4.4. Ablation

Here we evaluate our two main contributions: (1) the use of multi-modal information and joint modeling of both participants against vanilla self-attention (using only video and one participant at a time); and (2) the inclusion of longer-range temporal context ($T = 6$ and $T = 12$ chunk sequences, corresponding to 15.36 and 30.72 seconds, respectively) with respect to [58] ($T = 1$, i.e., 2.56 seconds). In order to mitigate the stochasticity introduced by the random initialization of the network weights, we repeated each experiment 4 times (or 8 for models with $T = 12$) and report the average of their results[2].

**Cross-modal and cross-subject attentions.** To assess the cross-attention's contribution we test four variants of our model: (1) a self-attention Transformer (TF$_v$) on the visual modality only and for each participant separately, i.e., attention is applied within each subject's sequence and neither cross-modal nor cross-subject attention are considered; (2) the Dyadformer with either only cross-modal attention (DF$_{xm}$) or (3) cross-subject attention (DF$_{xs}$); and (4) the full architecture with both cross-attentions (DF$_{xm,xs}$). As shown in Tab. 1, the two strongest variants were DF$_{xm}$ and DF$_{xm,xs}$. Although TF$_v$ was already a strong baseline model, it did not obtain the best result in any metric, suggesting that involving multiple modalities and explicitly modeling interaction among subjects is indeed beneficial for this task. The diminishing trend we observed on the performance of the models when further increasing their depth (number of encoder layers) discouraged us from trying further combinations and/or increasing their capacity with more parameters.

---
[2]Further ablations are included in the supplementary material.

**Temporal context.** We then evaluated different temporal context lengths, i.e., $T \in \{6, 12\}$, for the aforementioned combinations. As shown in Tab. 1, $T = 12$ achieves better results (lower MSE$_{seq}$ and MSE$_{part}$) throughout all the ablation. Interestingly, the Dyadformer variants with cross-subject attention, DF$_{xs}$ DF$_{xm,xs}$, benefited more from longer sequences. This is aligned to the fact that interpersonal dynamics can span very different temporal ranges. That is, the behavior of one interlocutor could be considerably delayed in time. Hence, using $T = 12$ allows such long-term interdependencies to emerge and be further leveraged.

**Use of metadata.** Preliminary experiments showed the benefits of their use at a marginal computational cost. Tab. 2 shows, for the simplest ablated model TF$_v$ ($L = 2$), that using only video results in very low values for the standard deviation. This *regression to the mean* problem is alleviated by allowing the model to access metadata information.

Note that the lack of metadata especially hurts *Extraversion* ("E"), *Agreeableness* ("A"), and *Negative emotionality* ("N"). If we compute the mean of the two sets of standard deviations (with and without metadata, from Tab. 2), we obtain 0.332 versus 0.116, respectively. This indicates the models are more willing to deviate the personality trait predictions from a mean value when incorporating the extra context provided by metadata. This is in line with current state-of-the-art research in personality psychology, which states that personality needs to be expressed in *situations* [62], i.e., taking the interaction context into account.

### 4.5. Analysis across personality traits and tasks

Here, we analyze the results obtained in the ablation studies described in Sec. 4.4. First, we evaluate the results from the four different tasks present in the UDIVA v0.5 dataset, as each of them was designed to elicit different behaviors. Then, we study how different tested variants of the Dyadformer model the different OCEAN traits, given that not all traits are equally expressed nor captured. We com-pare our results to the two best-performing models of [58]. Note that such models were trained per task, whereas our tested models were trained on all tasks jointly.

|  | O | C | E | A | N |
|---|---|---|---|---|---|
| Training (ground truth) | 0.255 ±1.136 | 0.160 ±1.020 | −0.053 ±0.969 | −0.006 ±0.957 | −0.346 ±1.085 |
| TF$_v$ ($L = 2$) wo/ metadata | −0.008 ±0.256 | 0.057 ±0.112 | −0.186 ±0.062 | −0.178 ±0.086 | −0.431 ±0.064 |
| TF$_v$ ($L = 2$) w/ metadata | −0.053 ±0.323 | 0.126 ±0.313 | −0.321 ±0.364 | −0.134 ±0.345 | −0.238 ±0.317 |

Table 2. Ablation on the regression to the mean problem. Mean and standard deviations of personality trait predictions by one run of the simpler TF$_v$ ($L = 2$) without and with metadata and the same values over the training ground truth for comparison.

**Per-task analysis.** As in [58], we analyzed the performance of the different model variations predicting the OCEAN traits separately depending on the task at hand. The results are shown in Tab. 3. As we can observe, among our models there is not a clear winner[3]. For *Animals*, $TF_v$ is the one which provided more accurate results on average ("Avg") both in terms of $MSE_{seq}$ and $MSE_{part}$, although $DF_{xm}$ did equally well for "A". $DF_{xs}$ outperformed the rest for the "N" trait in this task. Both for *Ghost* and *Lego*, $DF_{xm, xs}$ and $DF_{xm}$ got the lowest error in terms of $MSE_{seq}$ and $MSE_{part}$, respectively. Finally, for *Talk*, $DF_{xm, xs}$ outperformed the rest of the models on average, doing better than the rest for *Open-mindedness* ("O") and *Conscientiousness* ("C") measuring $MSE_{seq}$ and also for "O" and "E" measuring $MSE_{part}$ instead. Some of the findings diverge from the ones reported in [58]. For instance, whereas they found *Animals* to benefit more from audio than *Lego*, we see a contrary trend. However, note that our models are not trained in a task-specific fashion, thus the network has been able to learn from a wider range of behaviors encountered across tasks, which might impact the relative importance of each modality.

**Per-trait analysis.** Transversely to all tasks except for *Animals*, $DF_{xm, xs}$ is the most accurate model predicting "O" at participant-level. It is also the best at predicting "E" at participant-level and "C" at sequence-level, whereas $DF_{xm}$ does a better job at participant-level for the latter across all tasks. For "A", $DF_{xm, xs}$ is a close second after $DF_{xs}$. Interestingly, for "A", both variants incorporating cross-subject attention improved results. "A" is positively correlated with kindness, consideration, and cooperativeness, pro-social behaviors that are more clearly understood when the network attends to both interactants. In contrast, "N" does not usually benefit from cross-subject attention as this trait is more associated to the individual's inner context (i.e., stress, mood changes) [30]. Surprisingly though, we find opposite trends for *Animals*, for which "N" does highly

---

[3]Due to lack of space, Pearson correlation results (typically used in personality psychology [7]) are provided in the supplementary material to further validate our contributions.

benefit from cross-subject whereas "A" does not.

**Per-trait vs. per-task discussion.** While, on average, *Talk* is the task obtaining the lowest $MSE_{part}$ error, that is not the case per trait. If we focus on participant-level, the *Talk* scenario does allow to better predict "C", and "E", but *Animals* is more informative for "O" and "A", and *Lego* for "N". At sequence-level, "E" is better predicted with *Lego* and "N" with *Ghost*. These findings are consistent with those reported in [58]. This can be useful for psychological research, because it provides evidence that different situations actually enact different traits [20]. For the case of *Animals*, we can observe a strikingly low error for "A" followed by "O". This suggests that these two traits are likely enacted by this task. Trait-enactment refers to the idea that some situations enact, or activate, certain levels of traits required for this situation [19]. This pattern is confirmed when we look at *Talk*. Extraverted individuals are generally more talkative, but conscientious participants, even though they are not particularly extraverted, will engage in active talking when they are demanded to.

**Comparison to state of the art.** There exists only one previous work that published results on UDIVA v0.5 dataset [58]. The authors evaluated different model variants to complement the information from the "target person", the one whose personality was being predicted. Their simplest video-based model, namely "L", was enriched with either metadata ("m"), extended context ("E") – that is, the view from the other interlocutor – , and/or the audio ("a"). For their best model, namely "LEam", they reported a MSE$_{part}$ of 0.812, which is largely reduced by our best model by 12.5% (0.722 in Tab. 1). Moreover, for the different architecture alternatives, they report per-task/per-trait MSE$_{part}$. Interestingly, they noted that including the different sources of information could benefit or worsen the performance in different scenarios. Their two best-performing models were "LEm" and "LEam". The latter achieved generally better results, except for *Lego*, where the noise in the audio signal caused by the bricks might have hurt performance. Our different proposed models outperform their two best variations in 15/20 cases, as can be seen in Tab. 3.

## 5. Conclusion

We presented the Dyadformer, a novel multi-modal multi-subject transformer architecture for modeling individual and interpersonal features in dyadic interactions using variable time windows, thus allowing the capture of long-term interdependencies. We thoroughly ablate our model in the UDIVA v0.5 dataset for the task of self-reported personality prediction to demonstrate the contributions of each attentional module, as well as the modeling of longer timesteps. Experimental results demonstrated the reliability of our approach by surpassing previous results in UDIVA v0.5, reducing the error by 12.5% with respect to [58]. Re-sults also showed that context (or situations) matters in personality computing. Recently, situations have been put in the forefront of personality research to understand and predict real behavior [62]. In this sense, a promising extension of this work into the psychological realm would be to extract situational perceptions as we compute personality scores, since considering both features would undoubtedly improve behavior forecasting. In addition to audio/video-based personality computing, our model allows for straightforward adaptations to other modalities, as well as extend-ing our analysis to other individual and dyadic features. Future work will include the validation of the architecture for longer time windows, using other interaction datasets applied to different tasks, and exploring end-to-end learning which would allow for better coordination between backbones and the Dyadformer.

| Trait / Arch. | O | C | E | A | N | Avg |
|---|---|---|---|---|---|---|
| *Animals (A)* | | | | | | |
| [58] (LEm) | - / 0.736 | - / 0.834 | - / 0.968 | - / 0.669 | - / 1.192 | - / 0.880 |
| [58] (LEam) | - / 0.737 | - / **0.756** | - / **0.887** | - / 0.580 | - / 1.023 | - / 0.797 |
| TF$_v$ | **0.186** / 0.455 | 0.722 / 1.062 | **0.659** / 1.283 | **0.049** / **0.054** | 1.511 / 0.975 | **0.626** / **0.766** |
| DF$_{xm}$ | 0.206 / 0.515 | **0.691** / 1.008 | 0.677 / 1.328 | 0.050 / **0.054** | 1.658 / 1.041 | 0.656 / 0.789 |
| DF$_{xs}$ | 0.242 / 0.628 | 0.927 / 1.227 | 0.672 / 1.433 | 0.123 / 0.134 | **1.367** / **0.889** | 0.666 / 0.862 |
| DF$_{xm,xs}$ | 0.263 / 0.674 | 0.920 / 1.239 | 0.670 / 1.448 | 0.115 / 0.134 | 1.520 / 0.947 | 0.698 / 0.888 |
| *Ghost (G)* | | | | | | |
| [58] (LEm) | - / 0.743 | - / 0.944 | - / 0.868 | - / 0.657 | - / 1.153 | - / 0.873 |
| [58] (LEam) | - / **0.741** | - / 0.893 | - / 0.844 | - / 0.667 | - / 1.139 | - / 0.857 |
| TF$_v$ | 1.217 / 0.858 | 0.609 / 0.633 | 0.665 / 0.723 | 0.595 / **0.589** | 0.783 / **0.988** | 0.774 / 0.758 |
| DF$_{xm}$ | 1.231 / 0.889 | **0.563** / **0.584** | **0.629** / **0.707** | 0.615 / 0.617 | 0.778 / 0.989 | 0.763 / **0.757** |
| DF$_{xs}$ | 1.156 / 0.808 | 0.619 / 0.707 | 0.778 / 0.781 | **0.564** / 0.604 | 0.786 / 1.039 | 0.781 / 0.788 |
| DF$_{xm,xs}$ | **1.122** / 0.771 | 0.582 / 0.691 | 0.733 / 0.754 | 0.577 / 0.616 | **0.775** / 1.029 | **0.758** / 0.772 |
| *Lego (L)* | | | | | | |
| [58] (LEm) | - / **0.727** | - / 0.763 | - / 0.826 | - / **0.611** | - / 1.037 | - / 0.793 |
| [58] (LEam) | - / 0.745 | - / 0.839 | - / 0.953 | - / 0.659 | - / 1.099 | - / 0.859 |
| TF$_v$ | 0.925 / 0.808 | 0.806 / 0.657 | 0.514 / 0.755 | 0.614 / 0.710 | **0.534** / 0.866 | 0.679 / 0.759 |
| DF$_{xm}$ | 0.916 / 0.827 | 0.753 / **0.616** | **0.488** / 0.743 | 0.647 / 0.732 | 0.537 / **0.844** | 0.668 / **0.752** |
| DF$_{xs}$ | 0.847 / 0.749 | 0.801 / 0.663 | 0.575 / 0.789 | 0.555 / **0.709** | 0.567 / 0.975 | 0.669 / 0.777 |
| DF$_{xm,xs}$ | **0.808** / **0.741** | **0.727** / 0.635 | 0.517 / **0.736** | **0.527** / 0.747 | 0.555 / 0.908 | **0.627** / 0.753 |
| *Talk (T)* | | | | | | |
| [58] (LEm) | - / 0.825 | - / 0.718 | - / 0.878 | - / **0.639** | - / 1.047 | - / 0.821 |
| [58] (LEam) | - / 0.773 | - / 0.790 | - / 0.869 | - / 0.670 | - / **0.985** | - / 0.817 |
| TF$_v$ | 1.107 / 0.736 | 0.472 / 0.513 | 0.561 / 0.462 | 0.846 / 0.708 | 1.074 / **1.076** | 0.812 / 0.699 |
| DF$_{xm}$ | 1.117 / 0.735 | 0.467 / **0.488** | **0.526** / 0.440 | 0.862 / 0.719 | **1.057** / 1.081 | 0.806 / 0.693 |
| DF$_{xs}$ | 0.896 / 0.632 | 0.454 / 0.529 | 0.707 / 0.479 | **0.771** / **0.671** | 1.095 / 1.124 | 0.785 / 0.687 |
| DF$_{xm,xs}$ | **0.861** / **0.574** | **0.450** / 0.504 | 0.617 / **0.419** | 0.794 / 0.683 | 1.082 / 1.135 | **0.761** / **0.663** |

Table 3. Results per trait and task. For each model, first row is MSE$_{seq}$ and second row is MSE$_{part}$. The "Avg" column depicts the average performance per row (over all the traits). We compare to the best two models of [58]. Best result per task, trait, and metric in bold. Also, best result among our ablations underlined.

**Acknowledgments.** This work has been partially supported by the Spanish project PID2019-105093GB-I00 (MINECO/FEDER, UE) and CERCA Programme/Generalitat de Catalunya, and ICREA under the ICREA Academia programme.

# -Supplementary Material-

## 1. Additional ablation experiments

Here we include further experiments we performed to assess the validity of various design choices for the proposed Dyadformer. First, we evaluate an alternative design for the cross-attentional modules. Second, we explore the usefulness of the self-attentional modules at different stages of our model.

**Cross-attention versus bidirectional encoding.** Besides cross-attention, we also tried to follow the approach of bidirectional encoding from BERT [1] (discussed in Sec. 2 of the main paper). This alternative was implemented through two stages. First, two parallel multi-modal BERT encoders (which share weights among them and within them), each performing video-audio joint attention on its corresponding subjects. Then, their outputs are fed to a second stage with one BERT encoder, effectively attending over the two subjects. For a fair comparison with our $DF_{xm, xs}$ with $L_{xm}, L_{xs} \in \{1, 2\}$, we tried with different number of layers for the encoders of this BERT-like architecture such that the number of MHA blocks in both was similar. In particular, BERT with $L_{bm}, L_{bs}$, where $L_{bm}, L_{bs} \in \{3, 6\}$ are, respectively, the number of layers in the multi-modal BERT encoders and the multi-subject one. The BERT configuration $L_{bm} = L_{bs} = 3$ corresponds to the same number of attention layers included in our model with $L_{xm} = L_{xs} = 1$ and $L_{bm} = L_{bs} = 6$ corresponds to $L_{xm} = L_{xs} = 2$. Moreover, regardless of the combination of $(L_{bm}, L_{bs})$, the number of parameters of the architecture is 17.1M, which is comparable to either $DF_{xm}$ or $DF_{xs}$ (both with 19.4M). We set $T = 12$ for these experiments. We show the results on Tab. 1, where the other results are the same as the ones reported in Tab. 1 of the main paper. This variant resulted slightly worse than the equivalent Dyadformer variants ($DF_{xm,xs}$) for all metrics and combinations of layers. These results highlight the effectiveness of the used cross-attentional modules. One possible reason for this to happen is that our cross-attentional design helps decouple self-attention from accesses to the external memory (through separate MHA operations). The bidirectional encoding, however, emulates accesses to internal and external representations through a single multi-head attention, which may hinder learning to attend differently to one and the other.

| Arch. | L | | MSE$_{seq}$ | | MSE$_{part}$ | | Params |
|---|---|---|---|---|---|---|---|
| | | | $T=6$ | $T=12$ | $T=6$ | $T=12$ | |
| TF$_v$ | 2 | | 0.807 | 0.771 | 0.742 | 0.732 | 10.0M |
| | 4 | | 0.857 | 0.792 | 0.781 | 0.744 | |
| | 6 | | 0.919 | 0.856 | 0.837 | 0.807 | |
| | 8 | | 0.948 | 0.860 | 0.867 | 0.804 | |
| | $L_{xm}$ | $L_{xs}$ | $T=6$ | $T=12$ | $T=6$ | $T=12$ | |
| DF$_{xm}$ | 1 | - | **0.797** | 0.767 | **0.738** | 0.732 | 19.4M |
| | 2 | - | 0.845 | 0.767 | 0.777 | **0.722** | |
| | 3 | - | 0.880 | 0.802 | 0.824 | 0.762 | |
| DF$_{xs}$ | - | 1 | 0.802 | 0.768 | 0.763 | 0.745 | 19.4M |
| | - | 2 | 0.831 | 0.760 | 0.778 | 0.738 | |
| | - | 3 | 0.843 | 0.767 | 0.794 | 0.743 | |
| DF$_{xm,xs}$ | 1 | 1 | 0.831 | 0.760 | 0.794 | 0.741 | 36.0M |
| | 1 | 2 | 0.847 | 0.765 | 0.802 | 0.748 | |
| | 2 | 1 | 0.854 | **0.738** | 0.809 | **0.722** | |
| | 2 | 2 | 0.894 | 0.758 | 0.842 | 0.737 | |
| | $L_{bm}$ | $L_{bs}$ | $T=6$ | $T=12$ | $T=6$ | $T=12$ | |
| BERT | 3 | 3 | - | 0.818 | - | 0.784 | 17.1M |
| | 3 | 6 | - | 0.820 | - | 0.780 | |
| | 6 | 3 | - | 0.814 | - | 0.766 | |
| | 6 | 6 | - | 0.800 | - | 0.761 | |

Table 1. Ablation of different architectures and sequence lengths ($T$ chunks) in terms of average sequence- and participant-level mean squared errors: TF$_v$, a Transformer on each subject's sequence separately; DF$_{xm}$ or DF$_{xs}$, the Dyadformer with only cross-modal ("xm") or cross-subject ("xs") attention respectively; DF$_{xm,xs}$ with both; and BERT, an alternative for multi-modal multi-subject modeling. $L_.$ are the number of layers in the encoders. Best result per column in bold.

**Self-attention before cross-attention.** In preliminary experiments, the Dyadformer included self-attention modules before every cross-attention module. However, motivated by the observation of an overfitting trend for overly complex models, we considered discarding all self-attention modules so as to reduce the number of parameters.

As a result, for our model in Fig. 1 on the main document, we removed the self-attention encoder between the video embedding and the cross-modal encoder. The self-attention after the audio embeddings was kept to give the audio features a chance to evolve (as video embeddings do during the cross-modal attention), especially given the fact that audio embeddings were extracted from a model not fine-tuned on the personality prediction task – differently from video ones. Regarding the self-attention encoders prior to cross-subject encoders, we experimentally found the impact was negative when removing those layers in our best cross-subject models, i.e., DF$_{xs}$ and DF$_{xm, xs}$.

Without those layers, $MSE_{part}$ increases from 0.738 and 0.722 (reported in Tab. 1) to, respectively, 0.758 and 0.740.

## 2. Correlation analysis

In order to complement Tab.3 from the main text, we also report the Pearson correlation metric among the per-trait/per-task predictions and the self-reported personality ground truth for the participants in the test partition in Tab. 2.

By looking at this metric, $TF_v$ displayed the worst average ("Avg") results, mostly correlating negatively with the ground truth. A notable exception is, however, that it obtained the highest correlation (over 0.8) for the *Agreeableness* ("A") trait in *Animals* and *Ghost*.

In contrast, it can be observed that all of our Dyadformer variants correlated positively with the ground truth scores (except for $DF_{xm}$ in *Open-mindedness* ("O"), for which correlation is usually close to zero). $DF_{xm}$ was less accurate for *Conscientiousness* ("C"), *Extraversion* ("E") and *Negative emotionality* ("N") than $DF_{xs}$ when looking at the Pearson correlation, despite the opposite trend was observed looking at MSE-based metrics. $DF_{xs}$ correlated best with "N", although it showed poor correlation with "A" and "O".

$DF_{xm,xs}$ obtained the best "Avg" performance in terms of correlation for all the tasks, followed by $DF_{xs}$. This shows that explicitly modeling cross-subject interactions helps better approximate the distributions of the traits. The former achieved the highest correlation when predicting "O" and "E", even for *Animals*, where $MSE_{part}$ was very high. More concretely, its strongest correlations were found for the latter trait (~0.7). $DF_{xm,xs}$ was also the best correlating with "C", except for *Ghost*, where it ranked second. Nevertheless, and opposite to $DF_{xs}$, it correlated very poorly with "N", while obtaining reasonably good results in "A" for *Lego* and *Talk*.

## References

[1] Jacob Devlin, Ming-Wei Chang, Kenton Lee, and Kristina Toutanova. Bert: Pre-training of deep bidirectional transformers for language understanding. In *NAACL-HLT (1)*, pages 4171–4186, 2019. 2

| Arch. \ Trait | O | C | E | A | N | Avg |
|---|---|---|---|---|---|---|
| *Animals (A)* | | | | | | |
| $TF_v$ | **0.186** | 0.722 | **0.659** | **0.049** | 1.511 | **0.626** |
| | **0.455** | 1.062 | **1.283** | **0.054** | 0.975 | **0.766** |
| | -0.533 | 0.440 | -0.638 | **0.894** | 0.110 | 0.055 |
| $DF_{xm}$ | 0.206 | **0.691** | 0.677 | 0.050 | 1.658 | 0.656 |
| | 0.515 | **1.008** | 1.328 | **0.054** | 1.041 | 0.789 |
| | -0.020 | 0.524 | 0.458 | 0.406 | 0.339 | 0.342 |
| $DF_{xs}$ | 0.242 | 0.927 | 0.672 | 0.123 | **1.367** | 0.666 |
| | 0.628 | 1.227 | 1.433 | 0.134 | **0.889** | 0.862 |
| | 0.267 | 0.490 | 0.494 | 0.353 | **0.599** | 0.441 |
| $DF_{xm,xs}$ | 0.263 | 0.920 | 0.670 | 0.115 | 1.520 | 0.698 |
| | 0.674 | 1.239 | 1.448 | 0.134 | 0.947 | 0.888 |
| | **0.373** | **0.592** | **0.705** | 0.341 | 0.283 | **0.459** |
| *Ghost (G)* | | | | | | |
| $TF_v$ | 1.217 | 0.609 | 0.665 | 0.595 | 0.783 | 0.774 |
| | 0.858 | 0.633 | 0.723 | **0.589** | **0.988** | 0.758 |
| | -0.535 | **0.608** | -0.693 | **0.896** | 0.137 | 0.083 |
| $DF_{xm}$ | 1.231 | **0.563** | 0.629 | 0.615 | 0.778 | 0.763 |
| | 0.889 | **0.584** | **0.707** | 0.617 | 0.989 | **0.757** |
| | -0.028 | 0.565 | 0.470 | 0.387 | 0.343 | 0.347 |
| $DF_{xs}$ | 1.156 | 0.619 | 0.778 | **0.564** | 0.786 | 0.781 |
| | 0.808 | 0.707 | 0.781 | 0.604 | 1.039 | 0.788 |
| | 0.251 | 0.517 | 0.496 | 0.353 | **0.588** | 0.441 |
| $DF_{xm,xs}$ | **1.122** | 0.582 | 0.733 | 0.577 | **0.775** | 0.758 |
| | **0.771** | 0.691 | 0.754 | 0.616 | 1.029 | 0.772 |
| | 0.363 | 0.603 | **0.706** | 0.334 | 0.277 | **0.457** |
| *Lego (L)* | | | | | | |
| $TF_v$ | 0.925 | 0.806 | 0.514 | 0.614 | **0.534** | 0.679 |
| | 0.808 | 0.657 | 0.755 | 0.710 | 0.866 | 0.759 |
| | -0.588 | -0.042 | -0.741 | -0.212 | 0.193 | -0.278 |
| $DF_{xm}$ | 0.916 | 0.753 | **0.488** | 0.647 | 0.537 | 0.668 |
| | 0.827 | **0.616** | 0.743 | 0.732 | **0.844** | **0.752** |
| | 0.103 | 0.427 | 0.381 | 0.382 | 0.282 | 0.315 |
| $DF_{xs}$ | 0.847 | 0.801 | 0.575 | 0.555 | 0.567 | 0.669 |
| | 0.749 | 0.663 | 0.789 | **0.709** | 0.975 | 0.777 |
| | 0.351 | 0.495 | 0.512 | 0.354 | **0.511** | 0.445 |
| $DF_{xm,xs}$ | **0.808** | **0.727** | 0.517 | **0.527** | 0.555 | **0.627** |
| | **0.741** | 0.635 | **0.736** | 0.747 | 0.908 | 0.753 |
| | **0.510** | **0.580** | **0.714** | 0.388 | 0.215 | **0.481** |
| *Talk (T)* | | | | | | |
| $TF_v$ | 1.107 | 0.472 | 0.561 | 0.846 | 1.074 | 0.812 |
| | 0.736 | 0.513 | 0.462 | 0.708 | **1.076** | 0.699 |
| | -0.573 | 0.114 | -0.726 | -0.020 | 0.213 | -0.198 |
| $DF_{xm}$ | 1.117 | 0.467 | **0.526** | 0.862 | **1.057** | 0.806 |
| | 0.735 | **0.488** | 0.440 | 0.719 | 1.081 | 0.693 |
| | 0.193 | 0.452 | 0.419 | **0.404** | 0.312 | 0.356 |
| $DF_{xs}$ | 0.896 | 0.454 | 0.707 | **0.771** | 1.095 | 0.785 |
| | 0.632 | 0.529 | 0.479 | **0.671** | 1.124 | 0.687 |
| | 0.401 | 0.542 | 0.529 | 0.370 | **0.525** | 0.473 |
| $DF_{xm,xs}$ | **0.861** | **0.450** | 0.617 | 0.794 | 1.082 | **0.761** |
| | **0.574** | 0.504 | **0.419** | 0.683 | 1.135 | **0.663** |
| | **0.585** | **0.597** | **0.743** | 0.403 | 0.229 | **0.511** |

Table 2. Results per trait and task. For each model, first row is $MSE_{seq}$, second row is $MSE_{part}$, and third row is Pearson Correlation also at participant level (ranging in $[-1, 1]$, closer to 1 is better). The "Avg" column depicts the average performance per row (over all the traits). Best result per task, trait, and metric in bold.